\documentclass[conference,letterpaper]{IEEEtran}
\IEEEoverridecommandlockouts

\usepackage[T1]{fontenc}
\usepackage{cite}
\usepackage{amsmath,amssymb,amsfonts}
\usepackage{algorithm}
\usepackage{algpseudocode}
\usepackage{graphicx}
\usepackage[export]{adjustbox}
\usepackage{textcomp}
\usepackage{xcolor}
\usepackage{booktabs}
\usepackage{threeparttable}
\usepackage{stfloats}
\usepackage{hyperref}

\def\BibTeX{{\rm B\kern-.05em{\sc i\kern-.025em b}\kern-.08em
    T\kern-.1667em\lower.7ex\hbox{E}\kern-.125emX}}

\usepackage{forest}
\usepackage{tikz}
\usetikzlibrary{fit,positioning,shapes.geometric,calc}

\tikzset{
  red arrow/.style={
    midway,red,sloped,fill, minimum height=3cm, single arrow, single arrow head extend=.5cm, single arrow head indent=.25cm,xscale=0.3,yscale=0.15,
    allow upside down
  },
  black arrow/.style 2 args={-stealth, shorten >=#1, shorten <=#2},
  black arrow/.default={1mm}{1mm},
  tree box/.style={draw, rounded corners, inner sep=0.5em},
  node box/.style={white, draw=black, text=black, rectangle, rounded corners},
}

\title{Asymmetric Correspondence Learning for Lexical and Style Embeddings: Interpreting Latent Space via Acoustic Features}

\author{
  \IEEEauthorblockN{Michael Whealing, Robert Gaizauskas, and Thomas Hain}
  \IEEEauthorblockA{\textit{School of Computer Science}, \textit{University of Sheffield} \\
  \{mwhealing1, r.gaizauskas, t.hain\}@sheffield.ac.uk}
}

\begin{document}

\maketitle
% Suppress page number on title page
\thispagestyle{empty}

\begin{abstract}
    Self-supervised speech embeddings jointly encode lexical, speaker, and prosodic information, complicating feature isolation. We introduce Asymmetric Correspondence Learning (ACL), distilling HuBERT features into separate Acoustic Word Embeddings (AWEs) and Acoustic Style Embeddings (ASEs) via a split-latent correspondence bottleneck. Through probing and clustering evaluations on single-word utterances, we demonstrate that substantial variance from extrinsic sources remains encoded within each latent space, despite linear probing indicating the information has been stripped. Whilst the AWE space exhibits cohesive lexical clustering (AP$\approx$0.90, Fisher$\approx$6.28), linear probing on the ASE space achieves 97.9\% speaker accuracy across a diffuse topology (Fisher$\approx$1.01), yet preserves substantial lexical variance (Fisher$\approx$0.86) that remains linearly unrecoverable (0.17\%). To explain these findings, we apply group-level permutation feature importance over eGeMAPS acoustic features, finding that AWEs partition by vocal tract filters and ASEs by $F_0$ dynamics. These findings suggest that standard diagnostic benchmarks may be insufficient for verifying feature isolation, as substantial speech structure can be retained even when it remains entirely hidden from probes.
\end{abstract}

\section{Introduction}
Speech representations produced via Self-Supervised Learning (SSL) yield remarkable performance across downstream ASR, voice conversion, and speech synthesis tasks \cite{mohamed2022self}. Whilst isolating lexical, speaker, or prosodic elements into specialised latent spaces enhances performance and efficiency \cite{qu2023disentangling}, separating each feature is difficult due to natural speech variation across phonological, prosodic, and pragmatic contexts, which entangle within the representation \cite{jurafsky2003probabilistic}. To isolate lexical content, recent work leverages \textit{Acoustic Word Embeddings} (AWEs) to map variable-length sequences into fixed-dimensional vectors \cite{sanabria2023analyzing} that cluster identical words together \cite{meghanani-hain-2024-improving} while remaining invariant to speaker and prosodic changes \cite{lin2023selfsupervisedacousticwordembedding}.

Conventional AWE literature utilises latent pooling \cite{sanabria2023analyzing} or Correspondence Autoencoders (CAEs) \cite{DBLP:journals/corr/abs-1811-00403}, where non-lexical variance is cited as detrimental to downstream performance. These spaces are typically validated via word-retrieval and discrimination benchmarks \cite{abdullah2022integratingformmeaningmultitask, settle2016discriminativeacousticwordembeddings, kamper2016deepconvolutionalacousticword, algayres2020evaluatingreliabilityacousticspeech}. However, these probes are bounded by their sensitivity \cite{hewitt2019designinginterpretingprobescontrol}; a task succeeds if lexical content is dominant, regardless of remaining extrinsic info, which may be inaccessible to the probe, undermining downstream generalisation and speaker anonymisation.

To investigate this, we extend contemporary CAE frameworks \cite{meghanani-hain-2024-improving} using \textit{Asymmetric Correspondence Learning} (ACL), jointly learning an AWE for lexical content and an Acoustic Style Embedding (ASE) for stylistic content. Here we define \textbf{\textit{style}} as speech components modulating an utterance at the intersection of speaker identity and prosody \cite{zhu2024unistyle, SCHULLER20134}. This captures unique speaker signatures alongside variations that communicate intent or emotion \cite{DBLP:journals/corr/abs-1803-09017}.

While probing indicates isolation of word identity and speaker traits, unsupervised clustering reveals that substantial extrinsic variance remains encoded. To evaluate this structure, we apply a Random Forest classifier with group-level permutation feature importance \cite{breiman2001random} over eGeMAPS acoustic descriptors \cite{eyben2015geneva} to predict latent cluster pseudo-labels. We find that the AWE space partitions strictly along vocal tract filters (MFCCs, spectral bands), whereas the ASE coordinates smoothly along glottal source dynamics ($F_0$, spectral balance) associated with prosodic structure \cite{Triantafyllopoulos_2023}. This divergence suggests that standard diagnostic probes underestimate retained extrinsic information, showing that unrecovered traits remain highly structured along speech production dimensions that a probe may fail to capture.

The main contributions of this work are as follows:
\begin{enumerate}
    \item \textbf{Asymmetric Correspondence Learning (ACL):} A novel framework that factorises lexical and style traits into separate latent spaces without explicit supervision or orthogonality constraints (\S\ref{method: ACL}).
    \item \textbf{Evidence of Probing Limitations:} A case study demonstrating that near-floor linear probing performance does not guarantee feature erasure, providing evidence that extrinsic variance can remain embedded within latent speech representation spaces (\S\ref{res: baseline}).
    \item \textbf{Acoustic Interpretability:} An explainability method utilising group-level permutation feature importance to map unsupervised latent topologies directly back to speech production metrics (\S \ref{method: acoustic_analysis};\S\ref{sec: clusters}).
    \item \textbf{Latent Manifold Characterisation:} Detailed analysis demonstrating that the AWE structures along acoustic-phonetic properties while the ASE aligns along continuous prosodic dynamics (\S\ref{sec: clusters}).
\end{enumerate}

%The remainder of this paper is organised as follows. Section~\ref{sec:methods} details the ACL architecture and our permutation algorithm. Section~\ref{sec:implementation} outlines data preparation and model parameters. Section~\ref{sec:results} delivers downstream evaluations and baseline probing. Section~\ref{sec:clustering} evaluates emergent topologies via unsupervised clustering, and Section~\ref{sec:conclusion} concludes the paper.

\section{METHODS}
\subsection{Asymmetric Correspondence Learning}
\label{method: ACL}

\begin{figure}[H]
    \centering
    \includegraphics[width=1\linewidth]{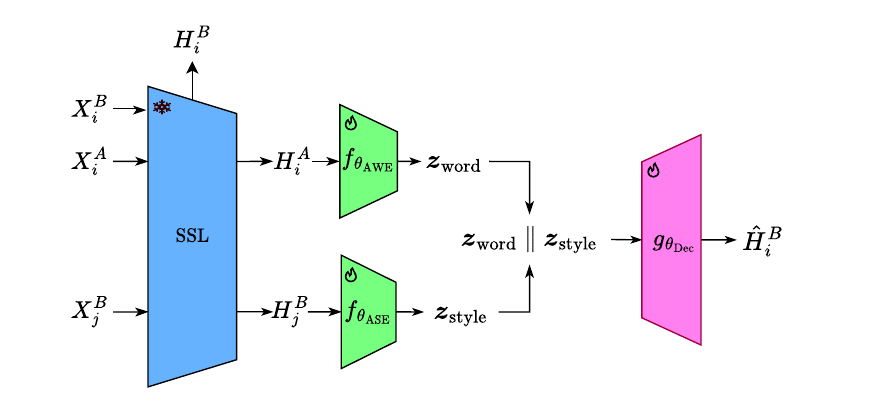}
    \caption{Pre-training scheme to extract AWE and ASE latent vectors using the ACL framework.}
    \label{method:model}
\end{figure}

Firstly, each element of a triplet of waveform utterances $\mathcal{T} = (X^{A}_{i}, X^{B}_{i},X^{B}_{j})$ is processed via a pre-trained base HuBERT \cite{DBLP:journals/corr/abs-2106-07447} model to extract high-dimensional acoustic features $H$ (see Figure \ref{method:model}). Here, $X^{A}_{i}$ denotes the waveform for speaker $A$ uttering the word $i$, $X^{B}_{i}$ denotes the waveform for speaker $B$ uttering the word $i$ and $X^{B}_{j}$ denotes the waveform for speaker $B$ uttering the word $j$, $j \neq i$.

\noindent The model consists of two architecturally identical but separately initialised encoders $f_{\theta}$ that produce latent representations that are then jointly fed into a decoder network. In this configuration, the encoder $f_{\theta_{AWE}}$ receives a lexical-positive sample from the triplet $\mathcal{T}$, while $f_{\theta_{ASE}}$ receives the style-positive (lexically negative) sample. The model is pre-trained to compress the input HuBERT features into latent vectors using an information bottleneck; see \ref{model:deets} for implementation details. Network $f_{\theta_{AWE}}$ generates the latent vector $z_{word}\in \mathbb{R}^d$ (AWE) and $f_{\theta_{ASE}}$ generates $z_{style}\in \mathbb{R}^d$ (ASE). These vectors are concatenated and passed to the decoder model, $g_{\theta_{DEC}}$, which produces its prediction of the acoustic feature vector for $B$ speaking word $i$, $\hat{H}_{i}^B$. This is then compared with the ``true" HuBERT encoding $H_{i}^B$ and the network is updated using a Mean Square Error (MSE) reconstruction loss, 
\begin{equation}
\label{eq: mse}
\mathcal{L}_{\text{MSE}} = \frac{1}{\sum^B_{b=1} M_b}
\sum_{b=1}^{B} M_b \cdot
\big( \hat{H}_{b,i}^B - H_{b,i}^B \big)^2
\end{equation}

\noindent Since SSL speech embeddings are variable length, mask $M_b$ is applied to each batch element $b$, to exclude padding frames from the loss calculation. This objective forces the model to synthesise the SSL features of Speaker $B$ (derived from the ASE) performing utterance $i$ (derived from the AWE).

\subsection{Cluster Interpretability via Group Feature Importance}
\label{method: acoustic_analysis}

Using the feature groups listed in Table \ref{tab:feat_groups} we independently analyse the  $K$-means clustered latent spaces of the ASE and AWE. For each space, a Random Forest (RF) classification model uses the acoustic feature groups to predict the pseudo cluster labels. The classifier identifies the primary acoustic drivers of the clusters using an adaptation of permutation feature importance \cite{breiman2001random} applied across the acoustic feature groups shown in Table \ref{tab:feat_groups}. This model-agnostic technique quantifies a group's structural contribution by measuring the degradation in macro-$F_1$ score when its constituent features are jointly altered using a shared permutation index, breaking their relationship with the target cluster assignments.

\begin{algorithm}[H]
\caption{Permutation Feature Group Importance}
\label{alg:pfgi}
\footnotesize{
\begin{algorithmic}[1]
\Require Trained surrogate model $\mathcal{M}$, validation dataset 
    $(X \in \mathbb{R}^{N \times D}, Y \in \{1,\dots,K\}^N)$, 
    feature groups $\mathcal{G}$, repetitions $R$, 
    smoothing factor $\epsilon = 10^{-9}$
\Ensure Sorted group importance metrics $\mathcal{G}_{\text{sorted}}$
\Statex
\State $\mathcal{F}_{\text{base}} \gets \text{Macro-}F_1\big(\mathcal{M}(X), Y\big)$
\For{each group $(G_{\text{name}}, G_{\text{feats}}) \in \mathcal{G}$}
    \State $\mathcal{I} \gets \{i \in \{1,\dots,D\} \mid \text{feature\_name}[i] \in G_{\text{feats}}\}$
    \If{$\mathcal{I} = \emptyset$}
        \State \textbf{continue}
    \EndIf
    \State $\mathcal{D} \gets [\,]$
    \For{$r = 1$ \textbf{to} $R$}
        \State $X_{\text{perm}} \gets \text{copy}(X)$
        \State $\pi \gets \text{permute}(\{1, \dots, N\})$
        \State $X_{\text{perm}}[n, \mathcal{I}] \gets X[\pi(n), \mathcal{I}] \quad \forall n \in \{1,\dots,N\}$
        \State $\delta^{(r)} \gets \mathcal{F}_{\text{base}} - \text{Macro-}F_1\big(\mathcal{M}(X_{\text{perm}}), Y\big)$
        \State Append $\delta^{(r)}$ to $\mathcal{D}$
    \EndFor
    \State $\mu_{\text{drop}} \gets \frac{1}{R}\sum_{r=1}^{R} \mathcal{D}[r]$
    \State $\sigma_{\text{drop}} \gets \sqrt{\frac{1}{R}\sum_{r=1}^{R} \big(\mathcal{D}[r] - \mu_{\text{drop}}\big)^2}$

    \State $\text{group\_importances}[G_{\text{name}}] \gets \{\mu_{\text{drop}},\, \sigma_{\text{drop}}\}$
\EndFor
\Statex
\State \Return $\mathcal{G}_{\text{sorted}} \gets \text{sort}\big(\text{group\_importances})$
\end{algorithmic}
}
\end{algorithm}

\noindent Covariance between features within the same group is preserved by generating a single permutation index, $\pi$, since independent shuffling creates unrealistic data points the RF model has not seen (e.g. simultaneous high and low pitch features). This alters the model's performance through destruction of the feature space. By grouping features and joint shuffling, feature space is preserved more accurately.

\section{IMPLEMENTATION DETAILS}

\subsection{Data Preparation}
\label{sec:data_prep}
All English single word utterances are selected from the Multilingual LibriSpeech (MLS) dataset \cite{Pratap2020MLSAL} , with meta-data containing word, speaker and duration labels supplied by \cite{meghanani-hain-2024-improving}.

\begin{table}[H]
    \centering
    \caption{A summary of statistics across each data split}
    \label{tab:data_stats}
    \resizebox{0.85\columnwidth}{!}{
    \begin{tabular}{l l l l}
        \toprule
        \textbf{Data Statistics} & \textbf{Train} & \textbf{Dev} & \textbf{Test} \\
        \midrule
        \# Spoken Words & 96,210 & 3,324 & 3,380 \\
        \# Unique Spoken Words & 7,881 & 2,547 & 2,645 \\
        \# Speakers & 180 & 42 & 41 \\
        \midrule
        Avg Utterance Duration (s) & 0.642 & 0.634 & 0.630 \\
        Total Duration (hours) & 17.16 & 0.59 & 0.59 \\
        \bottomrule
    \end{tabular}%
    }
\end{table}
\noindent Data was filtered in line with standard AWE literature \cite{DBLP:journals/corr/abs-1811-00403}, where utterances with a duration equivalent or greater than 0.5 seconds and frequency between 5 and 50 in the training set are accepted as valid vocabulary. Speaker identities across splits are non-overlapping, which is desirable for both AWE (robust to speaker variation) and ASE (zero-shot speaker identification) evaluation. The MLS dataset is  derived from read audiobooks from LibriVox. Whilst audiobooks offer high acoustic quality and speaker specific stylistic variation, we acknowledge the limitations of read speech for capturing natural prosody. Evaluation of the framework on spontaneous conversational corpora remains an open direction for future work. 

%\subsection{SSL Feature Extraction}
%Pre-trained SSL models are trained on large amounts of unlabelled speech corpora. Whilst these models are able to capture rich acoustic information, the resulting representations are heavily contextualised within the global utterance \cite{mohamed2022self}. Consequently, word-level features derived from labelled timestamps retain residual information from neighbouring segments, making it difficult to isolate pure lexical content from the surrounding stylistic and temporal context.

\subsection{Triplet Generation}
\label{sub_sec:tri_gen}

An advantage of the triplet generation scheme is the scalability of the dataset. The general expression for exhaustive triplet generation can be expressed roughly through, 
\[
    |\mathcal{T}| \approx N_S(N_S - 1) \times |V|(|V| - 1) \times \bar{C}^3
\] 
\noindent where variable $N_S$ denotes the number of speakers, $|V|$ the vocabulary, and $\bar{C}$ the average number of instances per word, per speaker. The size of the dataset increases quadratically as more unique speakers and vocabulary are included, $|\mathcal{T}| \approx \mathcal{O}(N^2_S \cdot |V|^2)$. For each split we generate a sample pool of $10^8$ triplets - equivalent to approximately 45,000 hours of unique triplet combinations - from which we derive smaller subsets for experimentation (shown in Table \ref{tab:data_dist}). Given an average segment duration of $0.6$ seconds, each triplet effectively exposes the model to $1.8$ seconds of acoustic context.

\begin{table}[H]
    \centering
    \resizebox{0.6\linewidth}{!}{
    \begin{threeparttable}
        \caption{Experimental Dataset Summary}
        \label{tab:data_dist}
        
        \begin{tabular}{l l l}
            \toprule
            \textbf{Split} & \textbf{Triplets} & \textbf{Duration (Hr)} \\ 
            \midrule
            \textbf{Train} & 500,000 & $\approx 250$ \\
            \textbf{Val.}  & 5,000   & $\approx 2.5$ \\
            \textbf{Test}  & 20,000   & $\approx 10$ \\
            \bottomrule
        \end{tabular}
    \end{threeparttable}%
    }
\end{table}

\subsection{Acoustic Feature Groups}
\label{sub:feat_groups}

We investigate the AWE and ASE spaces by grouping related acoustic features taken form the eGeMAPS feature set as outlined in Table \ref{tab:feat_groups}. Grouping decisions are grounded in source-filter theory \cite{fant1981source}, which characterises speech production as a two-stage process where sound is generated by a glottal source, which is subsequently shaped by the resonant properties of the vocal tract \cite{10.1093/acrefore/9780199384655.013.894, jm3}. This decomposition naturally separates prosodic features (source) from acoustic-phonetic features (filter). Our hypothesis is that grouping features into distinct acoustic domains (e.g., aggregating all $f_0$ metrics or all spectral balance features together) provides a more reliable interpretation of the latent spaces than permuting individual features in isolation. This domain-level approach maintains the integrity of the evaluation by preserving intra-domain collinearities and minimising physically impossible feature combinations. While group permutation reduces the unrealistic distortions caused by single-feature permutation, we acknowledge this setup does not capture subtler cross-group correlations (such as the various formants).

\begin{table}[H]
\centering
\caption{Acoustic feature groups derived from eGeMAPS used in the interpretability analysis.}
\label{tab:feat_groups}
\resizebox{0.9\linewidth}{!}{
\begin{tabular}{l l p{3cm}}
\toprule
\textbf{Group} & \textbf{Property} & \textbf{Description} \\
\midrule
F0 Prosody      & Pitch              & F0 contour, intonation, prosodic movement. \\
Jitter          & Pitch Perturbation & Cycle-to-cycle F0 variation. \\
Formant F1      & Resonance (F1)     & Vowel height and jaw aperture. \\
Formant F2      & Resonance (F2)     & Vowel frontness, tongue position. \\
Formant F3      & Resonance (F3)     & Individual vocal tract shape. \\
Loudness        & Intensity          & Loudness dynamics, speaking effort. \\
Shimmer         & Amplitude Perturbation    & Cycle-to-cycle amplitude variation. \\
HNR             & Glottal Noise      & Harmonics-to-noise ratio; breathiness. \\
Spectral Balance & Spectral Energy   & Spectral slopes, flux, harmonic diff. \\
MFCC            & Cepstral Shape     & Vocal tract filter envelope. \\
Temporal        & Rhythm             & Voiced/unvoiced timing, syllable rate. \\
\bottomrule
\end{tabular}
}
\end{table}

\subsection{Model Details}
\label{model:deets}

\subsubsection{ACL Framework} The model (Fig \ref{method:model}) employs a Siamese dual-encoder architecture to process final layer, pre-extracted HuBERT base features (764-dim). The two encoders compress input feature frames into latent vectors. These latents are layer normalised \cite{ba2016layernormalization} and concatenated, forming a 256-dimensional vector, serving as the input for the decoder network. Each network is implemented as 2-layer Bidirectional GRU with 128 hidden dimensions and GeLU \cite{DBLP:journals/corr/HendrycksG16} activations. We apply a 0.1 dropout rate across all layers and train for 20 epochs using the MSE reconstruction loss (eq. \ref{eq: mse}) with a learning rate of 1.5e-5, weight decay of 0.01, and a batch size of 64.

\subsubsection{Interpretability Framework} A RF classifier, configured with 250 trees and balanced class weightings, was trained on an 80/20 train-test split of the 88-dimensional eGeMAPS feature vectors extracted from each audio sample in the triplet dataset. The classifier was tasked with predicting pseudo-cluster labels, which were derived via unsupervised $K$-means clustering of the latent embedding spaces produced by ACL inference on the test set. To validate the model, Out-of-bag estimates \cite{breiman1996out} are compared against dummy baselines. Evaluation on the unperturbed validation data established a reference score ($\mathcal{F}_{\text{base}}$) prior to feature shuffling. Finally, using Algorithm \ref{alg:pfgi}, each group's permutation routine was executed for $R=5$ repetitions to determine the feature group importances.

\vspace{-0.25cm}
\section{RESULTS}
\subsection{Baseline Performance}
\label{res: baseline}
We evaluate the embedding spaces of $z_{\text{word}}$ and $z_{\text{style}}$ using AWE retrieval and discrimination tasks, comparing them against conventional Correspondence Autoencoder (CAE) networks for baseline comparisons. The baseline AWE network utilises an autoencoder bottleneck that receives $H^A_i$ to generate the latent $z_{\text{word}}$, and attempts to reconstruct $H^B_i$ without $z_{\text{style}}$ concatenation. Conversely, the baseline ASE network swaps the input from the lexically positive sample $H^A_i$ to a style sample $H^B_j$, while still decoding into $H^B_i$. To evaluate each representation space, we employ four complementary metrics. 

Firstly, we evaluate downstream instance retrieval using macro-averaged Average Precision (AP), which computes the area under the precision-recall curve to measure how effectively matching target classes are ranked by distance. Then, to measure linear recoverability, a Logistic Regression model is trained on an 80/20 split of the evaluation dataset. To rigorously test generalisation and prevent category leakage, we ensure target class labels are non-overlapping across the train and test splits; for example, when probing for lexical content (word labels), the speaker identities in the training split do not overlap with those in the test split, and vice versa when probing for speaker identity. Downstream capability is quantified as linear recoverability, which we measure via the classification accuracy of a multinomial logistic regression probe. Performance on the target latent attribute is reported as classification accuracy, while the probe's accuracy on the opposing attribute is reported as leakage accuracy. Additionally, to determine the separability and discriminative power of the latents, we calculate the ratio of inter-class variance to intra-class variance for both speaker and word labels,

\vspace{-0.5cm}
    \begin{equation}
    \label{eq:fisher}
        \text{FisherScore}(\boldsymbol{Z}) = \frac{\sum^{|C|}_{c=1}n_c||\boldsymbol{\mu}^{(c)} - \boldsymbol{\mu}||^2_2}{\sum^{|C|}_{c=1}\sum_{i \in c}||\boldsymbol{z}_i - \boldsymbol{\mu}^{(c)}||^2_2}
    \end{equation} 
\vspace{-0.5cm}

\noindent where $n_c$ is the number of samples in class $c$, $\boldsymbol{\mu}^{(c)}$ is the class centroid, and $\boldsymbol{\mu}$ represents the global mean of the latent space $\boldsymbol{Z}$. Standard Fisher scores evaluate the primary target label of the space, while Fisher Leakage measures the space against its antithetical label. Finally, while F-scores quantify variance relative to class centroids, we implement a non-parametric, point-to-point metric, $R$, to evaluate local neighbour class configurations,
\vspace{-0.5cm}
    \begin{equation}
        \label{eq:intra_inter_ratio}
        R = \frac{\mathcal{D}_{\text{intra}}}{\mathcal{D}_{\text{inter}}} = \frac{\displaystyle \frac{1}{|Y|} \sum_{c \in Y} \frac{1}{\binom{N_c}{2}} \sum_{i < j} d_{\text{cos}}\left(\mathbf{z}_i^{(c)}, \mathbf{z}_j^{(c)}\right)}{\displaystyle \frac{1}{\binom{|Y|}{2}} \sum_{c \neq c'} d_{\text{cos}}\left(\boldsymbol{\mu}^{(c)}, \boldsymbol{\mu}^{(c')}\right)}
    \end{equation}
\vspace{-0.5cm}
    
\noindent where $Y$ is the set of unique target categories (words or speakers), $N_c$ is the number of samples in class $c$, $\mathbf{z}_i^{(c)}$ represents the embedding vectors, $\boldsymbol{\mu}^{(c)}$ is the class centroid, and $d_{\text{cos}}$ is the cosine distance. Crossing the ``unity'' boundary ($R < 1$) indicates well-separated compact clusters, whereas $R > 1$ signifies a diffuse, overlapping distribution where instances are, on average, closer to out-of-class points than to their own class cohorts. Table~\ref{tab:combined_analysis} presents these comparative results of the ACL against baseline CAE networks.

\begin{table}[H]
\centering
\caption{Downstream Probing and Manifold Analysis}
\label{tab:combined_analysis}
\setlength{\tabcolsep}{2.9pt}
\begin{tabular}{lcccc}
\toprule
 & \multicolumn{2}{c}{\textbf{AWE (Word Space)}} & \multicolumn{2}{c}{\textbf{ASE (Style Space)}} \\
\cmidrule(lr){2-3} \cmidrule(lr){4-5}
\textbf{Evaluation Metric} & \textbf{Baseline} & \textbf{ACL} & \textbf{Baseline} & \textbf{ACL} \\
\midrule
Fisher Score ($\uparrow$)       & 5.61    & $\boldsymbol{6.28}$    & 0.77    & $\boldsymbol{1.01}$    \\
Classification Acc. ($\uparrow$) & 82.1\% & $\boldsymbol{84.0\%}$ & 97.9\% & $\boldsymbol{97.9\%}$ \\
Average Precision ($\uparrow$)    & 0.89   & $\boldsymbol{0.90}$   & 0.17   & $\boldsymbol{0.18}$   \\
\midrule
Fisher Leakage ($\downarrow$)    & 0.083   & 0.083   & 0.97  & $\boldsymbol{0.86}$   \\
Leakage Accuracy ($\downarrow$)  & $\boldsymbol{2.89\%}$ & 3.97\%   & 0.21\%  & $\boldsymbol{0.17\%}$  \\
\midrule
Pairwise Dist. (R)($\downarrow$) & 0.15 & $\boldsymbol{0.13}$ & 1.14 & $\boldsymbol{0.84}$ \\
\bottomrule
\end{tabular}
\end{table}

\subsection{Discussion}

\noindent Table \ref{tab:combined_analysis} indicates that ACL induces better latent attribute separation compared to the CAE baselines. The AWE space is partitioned by lexical content, achieving high word classification accuracy ($84.0\%$), retrieval ($\text{AP}=0.90$), and cluster cohesion ($\text{Fisher}=6.28$), while successfully suppressing speaker details to near-floor leakage ($3.97\%$). 

Conversely, the ASE space yields high speaker recoverability ($97.9\%$) within a more diffuse topology ($\text{Fisher}=1.01$). The marginal difference between the ASE speaker Fisher score ($1.01$) and its lexical leakage score ($0.86$), paired with floor word retrieval ($\text{AP}=0.18$), demonstrates that speaker identity is encoded continuously across the manifold rather than via discrete clusters. These structural shifts suggest that the collaborative joint-decoding bottleneck distributes feature variance based on the specific data attributes preserved between the network's input and output. We hypothesise that this cooperative routing stretches variance along the primary acoustic axis exposed to the encoder, while compressing variance along dimensions corresponding to structural properties handled by the parallel stream.

The structural shifts in Table \ref{tab:combined_analysis} reflect this variance redistribution. Lacking cooperative optimisation, the baseline ASE CAE must independently minimise reconstruction loss, forcing its variance across uncoordinated acoustic dimensions to capture local phonetic and sequence-level lexical dynamics alongside speaker traits. This results in higher baseline Fisher leakage ($\approx0.97$) and a diffuse topology with a pairwise distance ratio above unity ($\approx1.14$). 

Under ACL, the joint-decoding bottleneck distributes optimisation labour. Because the parallel lexical encoder ($f_{\theta_{\text{AWE}}}$) efficiently processes local sequence characteristics, $f_{\theta_{\text{ASE}}}$ is freed from maintaining explicit linear alignments for lexical content. Instead, extrinsic lexical variance is compressed into a subspace that is inaccessible to the probe (distance ratio $\approx0.84$, Fisher score $\approx0.86$), even as the space maintains a $97.9\%$ speaker classification accuracy.

\section{UNSUPERVISED STYLE CLUSTERING}
\label{sec: clusters}

\subsection{Emergent Clusters}

%he $K$-means algorithm partitions the latent space based on geometric similarity, allowing us to evaluate how cluster boundaries align with word or speaker identities across varying granularities ($k$) \cite{kamper2017embedded}

Since discrete word or speaker labels cannot fully describe the structure of the ASE space, we attempt to analyse the space through the application of $K$-means clustering. By calculating the Adjusted Mutual Information (AMI) between the cluster assignments and ground-truth labels, we may quantify and cross-verify the shared structural information within each embedding space. For these evaluations, $K$ is set to the unique vocabulary size ($|V|$) for words, and the total number of speakers ($N_S$) for speaker labels.
\vspace{-0.5cm}

\begin{table}[H]
    \centering
    \caption{Adjusted Mutual Information Analysis}
    \label{tab:clustering_metrics}
    \small
    \begin{tabular}{lcc}
        \toprule
        \textbf{Embedding Space} & \textbf{Word Labels} & \textbf{Speaker Labels} \\
        \midrule
        AWE & 0.94 & 0.29 \\
        ASE & 0.24 & 0.44 \\  
        \bottomrule
    \end{tabular}
\end{table}

\noindent Table \ref{tab:clustering_metrics} indicates the AWE isolates lexical structure, maintaining a $0.94$ AMI for word labels while restricting speaker traits. However, within the ASE space, while speaker identity AMI ($0.44$) exceeds word label AMI ($0.24$), speaker identity as an organising factor is drastically reduced compared to the strict lexical coherence of the AWE space. This structural footprint directly contrasts with the total dichotomy reported by linear probes ($97.9\%$ speaker versus $0.17\%$ word). This divergence demonstrates that diagnostic probes fail to account for the true distribution of latent information; the correspondence bottleneck is far less lossy than probing suggests. Rather than completely erasing extrinsic attributes, the embedding spaces preserves them non-linearly. 

To attempt to better understand the probing and AMI results, we apply Algorithm \ref{alg:pfgi} over the acoustic feature groups in Table~\ref{tab:feat_groups} to predict $K$-means pseudo-labels across varying $k$, identifying which acoustic features drive cluster partitioning. To enable scale-invariant comparison, we normalise the Permutation Feature Group Importance (nPFGI) scores various $k$, expressing a specific group importance, $I_G$, as a relative proportion of the absolute degradation observed across the entire set of feature groups $\mathcal{G}$, \begin{equation}
    \label{eq:npfgi}
     nPFGI= \frac{I_G}{\sum_{G' \in \mathcal{G}} |I_{G'}|}
\end{equation}

\noindent A higher nPFGI value indicates that a greater proportion of cluster assignments are driven by that specific acoustic group.
\vspace{-0.5cm}

\begin{figure}[H]
    \centering
    \caption{ASE nPFGI Scores}
    \label{fig:ase_gdi}
    \smallskip
    
    % 1. Include the full, untrimmed image
    \includegraphics[width=0.85\linewidth, valign=m]{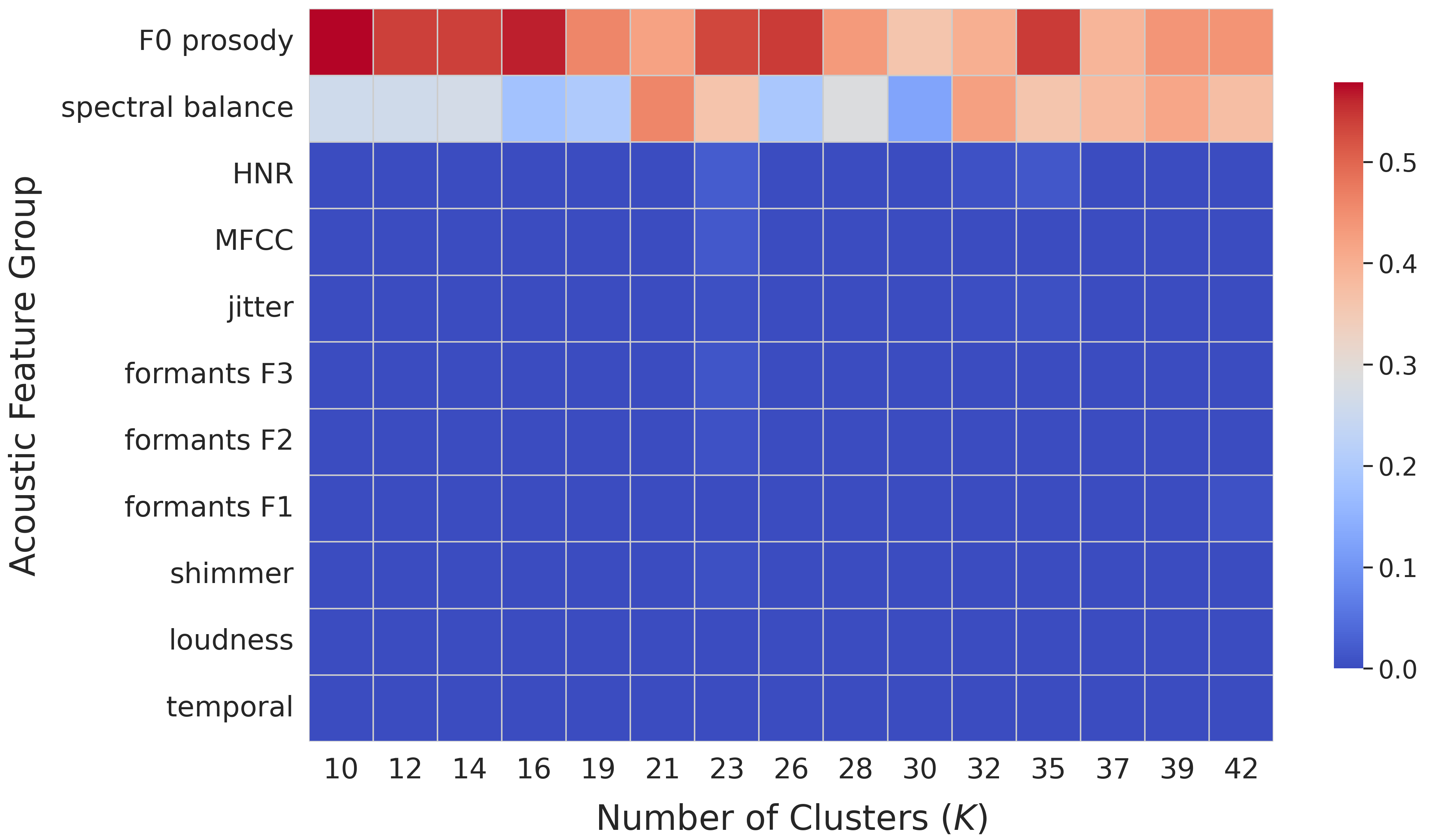}%
\end{figure}

\noindent Figure \ref{fig:ase_gdi} shows that $F0$ prosody features consistently dominate cluster classification across all $k$, with minor secondary contributions from spectral balance. Remaining feature groups contribute negligibly to pseudo-label classification. As $k$ approaches the number of speakers ($N_S=42$), $F0$ and spectral balance yield lower nPFGI scores and converge. These observations suggest that at lower $k$ granularities, the ASE space is primarily partitioned by $F0$ prosodic dynamics, consistent with the space organising around glottal source properties rather than discrete speaker or lexical categories.
\vspace{-0.25cm}
\begin{figure}[H]
    \centering
    \caption{AWE nPFGI Scores}
    \label{fig:awe_gdi}
    \smallskip
    \includegraphics[width=0.85\linewidth]{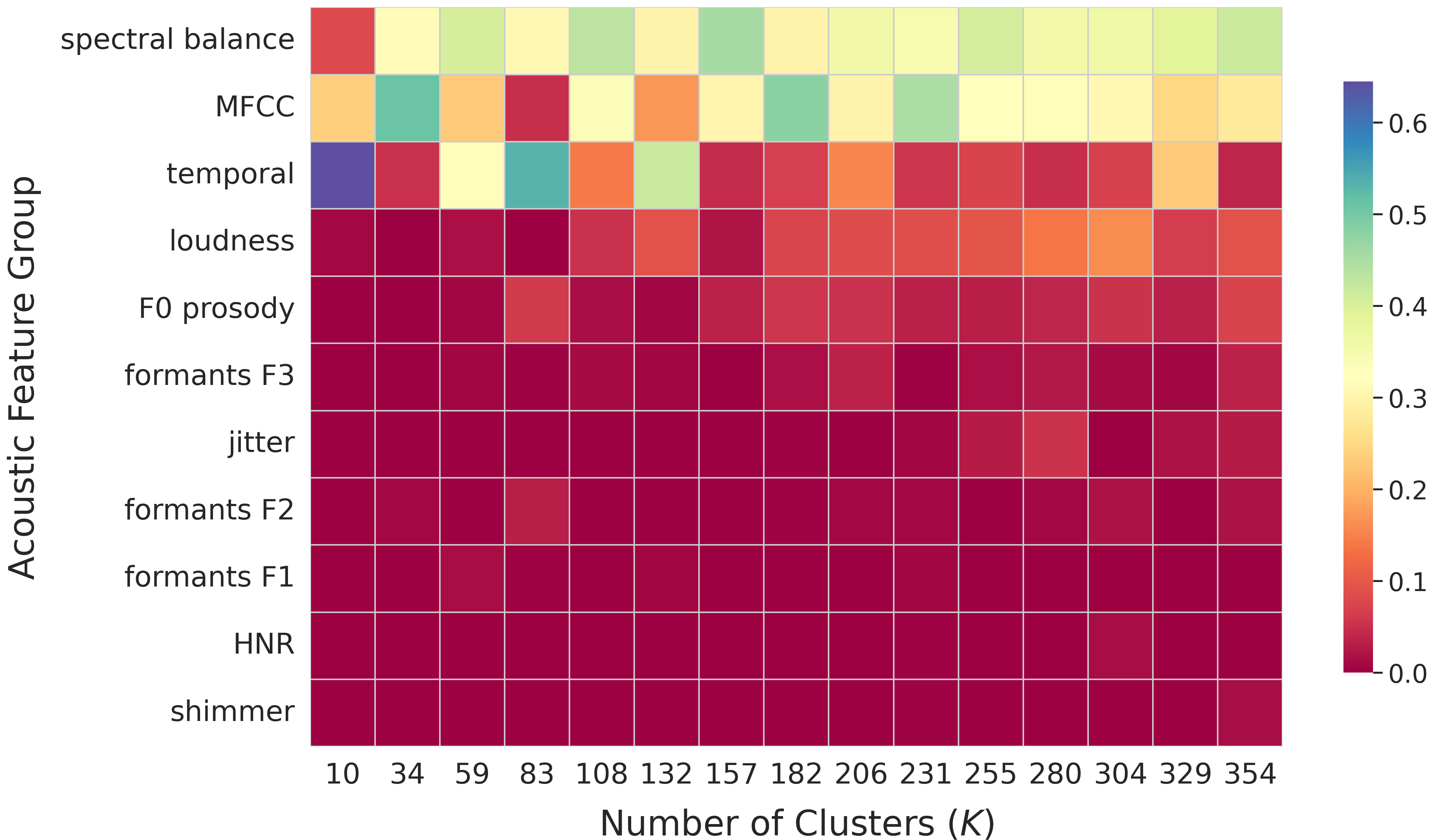}
\end{figure}
\noindent Figure \ref{fig:awe_gdi} shows spectral balance and MFCC features as the primary drivers of AWE cluster classification, with temporal features contributing more notably at lower $k$. The $F0$ feature group contributes negligibly throughout. Notably, the overall nPFGI magnitudes are lower than those observed in the ASE, suggesting that AWE cluster structure is less strongly determined by any single acoustic group compared to the ASE's dependence on $F0$. Between the two spaces, while they share some spectral importances, the spaces present juxtaposed feature groups to decide cluster assignments. The ASE relies on $F0$ features, while the AWE relies on MFCC and spectral features, with temporal elements being important at low $k$.

\subsection{Visualising Cluster Characteristics}

We compare pairs of utterances featuring the same speaker and word across different ASE clusters by mapping their acoustic profiles alongside corresponding spectrograms. At $k$ granularity, the radar axes plot the six acoustic features that most impact the nPFGI scores, chosen based on the magnitude of their cluster-specific $z$-scores. These profiles represent cluster-wide averages and are evaluated against a global baseline computed across all clusters in the system (dotted polygon).

\begin{figure*}[!t] 
    \centering
    \includegraphics[width=0.73\linewidth]{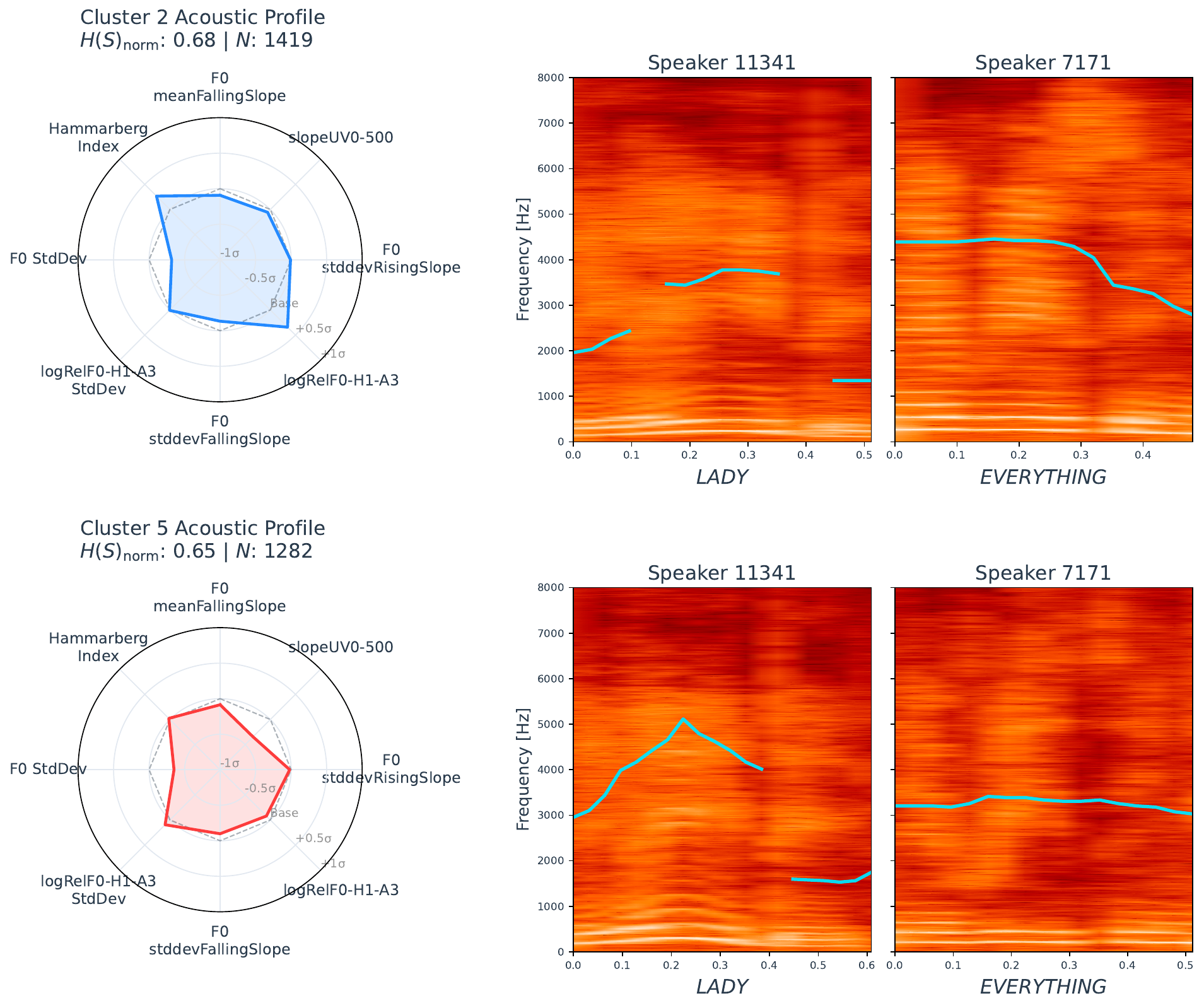} 
    \caption{Clusters generated using $k=14$. $H(S)_{\text{norm}}$ and $N$ denote normalised speaker entropy and number of data points in the cluster respectively; the cyan line tracks $F0$.}
    \label{fig:cluster_specs}
\end{figure*}

Figure \ref{fig:cluster_specs} shows Cluster 2 favours a continuous aperiodic noise style, visually evident in the upper spectrogram regions (above 4000 Hz) where turbulent glottal airflow creates an unvoiced noise band \cite{ogden2023introduction}. This aperiodicity interrupts vertical voicing striations, causing tracking breaks in the $F0$ contours. This is corroborated by positive cluster $z$-scores for the Hammarberg Index and logRelF0-H1-A3, implying a breathy speaking style. Conversely, Cluster 5 exhibits a periodic, modal speech style. Its voice quality metrics track the baseline, displaying cleaner harmonic structures and regularly repeating waveforms. Cluster 5 opposes Cluster 2 with a negative $z$-score in slopeUV0-500, indicating sharper attenuation of transient and continuous aperiodic energy during voiced segments \cite{alipour2012measures}, manifesting as compact intensity envelopes. Visual consistency across spectrograms verifies that the ASE space organises by structural differences in speech production. Furthermore, high normalised speaker entropy ($H(S)_{\text{norm}}$) suggests the space partitions into distinct prosodic styles rather than degenerating into speaker identity clusters.

\vspace{-0.15cm}
\section{Conclusion}

This work introduces Asymmetric Correspondence Learning (ACL), a framework designed to distil HuBERT features into separate Acoustic Word and Style Embeddings without explicit supervision or orthogonality constraints. Evaluation reveals a clear discrepancy between standard diagnostic metrics and the underlying latent structure. While probing implies successful isolation of lexical content and speaker identities, unsupervised clustering and AMI analysis suggest that substantial extrinsic speech variance remains embedded within both spaces. To explore the factors driving this latent organisation, an RF classifier (Algorithm \ref{alg:pfgi}) was trained to predict unsupervised cluster pseudo-labels using eGeMAPS acoustic features of our reconstruction target ($H_i^B$) utterances. Permutation group feature importance analysis indicates that AWE clusters are heavily influenced by vocal tract descriptors (such as MFCCs and spectral bands), whereas ASE clusters are driven by glottal source dynamics ($F_0$ prosody and spectral balance). Crucially, our core thesis extends beyond the known sensitivity gap between linear and non-linear probes \cite{hewitt2019designinginterpretingprobescontrol}; the key insight is that this unrecovered variance is not arbitrary noise, but is systematically structured along physical speech production dimensions that correspond directly to source-filter theory \cite{fant1981source, 10.1093/acrefore/9780199384655.013.894}. Standard disentanglement benchmarks are therefore insufficient not merely because non-linear probes recover more than linear ones, but because the retained structure is acoustically grounded in ways that neither probe type is designed to detect.

\vspace{-0.1cm}

These findings highlight a potential limitation in speech disentanglement, particularly in the case of separating the content from speaker. Because physical speech production intrinsically entangles speaker traits with both vocal tract geometry and glottal dynamics, future work aims to pivot from abstract categorical supervision toward physically grounded acoustic objectives, such as explicit source-filter factorisations. Such representations will be evaluated using standard downstream applications - including zero-shot voice conversion, speaker anonymisation, and prosody transfer - assessing whether source-filter aligned bottlenecks provide superior synthesis control and identity privacy in real-world settings. Given that the ASE space encodes lexical variance in a form undetectable by standard probes, speaker anonymisation applications warrant particular scrutiny, as identity-adjacent information may persist in ways current evaluation pipelines would fail to flag.

\section{Acknowledgements}

\noindent This work was supported by the industrial sponsorship associated with the UKRI AI Centre for Doctoral Training in Speech and Language Technologies (SLT) and their Applications which is funded by UK Research and Innovation [grant number EP/S023062/1].

\smallskip

\noindent During the preparation of this manuscript, the authors utilised Gemini (a large language model developed by Google) for technical advice regarding software implementation, structural feedback on the figure and table layouts, general grammar, and spell checks across all sections. No original text or experimental data was generated by the AI system, and the final manuscript was fully reviewed and verified by the authors.

\bibliographystyle{IEEEtran}
\bibliography{sample}

\end{document}